\documentclass[runningheads]{llncs}
\usepackage{amsmath}
\usepackage{amsfonts}
\usepackage{amsmath} 
\usepackage{placeins}
\usepackage{float}
\usepackage{hyperref}
\usepackage[T1]{fontenc}
\usepackage{newtxtext}
\usepackage{newtxmath}
\usepackage{graphicx}
\usepackage[
backend=biber,
]{biblatex}
\begin{document}
\title{AdURA-Net: Adaptive Uncertainty and Region-Aware Network}
\author{Antik Aich Roy\inst{1}\orcidID{0009-0003-0572-2424} \and
Ujjwal Bhattacharya\inst{2}\orcidID{0000-0002-8546-6453}} 
\authorrunning{A. Aich Roy and U. Bhattacharya}
\institute{Indian Statistical Institute, Kolkata, India \\
\email{antikaichroy\_t@isical.ac.in, ujjwal@isical.ac.in}}
\maketitle              
\begin{abstract}
One of the common issues in clinical decision-making is the presence of uncertainty, which often arises due to ambiguity in radiology reports, which often reflect genuine diagnostic uncertainty or limitations of automated label extraction in various complex cases. Especially the case of multilabel datasets such as CheXpert, MIMIC-CXR, etc., which contain labels such as positive, negative, and uncertain. In clinical decision-making, the uncertain label plays a tricky role as the model should not be forced to provide a confident prediction in the absence of sufficient evidence. The ability of the model to say it does not understand whenever it is not confident is crucial, especially in the cases of clinical decision-making involving high risks. Here, we propose AdURA-Net, a geometry-driven adaptive uncertainty-aware framework for reliable thoracic disease classification. The key highlights of the proposed model are: a) Adaptive dilated convolution and multiscale deformable alignment coupled with the backbone Densenet architecture capturing the anatomical complexities of the medical images, and b) Dual Head Loss, which combines masked binary cross entropy with logit and a Dirichlet evidential learning objective.

\end{abstract}
\begin{figure}[t]
    \centering
    \small
    \includegraphics[width=\linewidth]{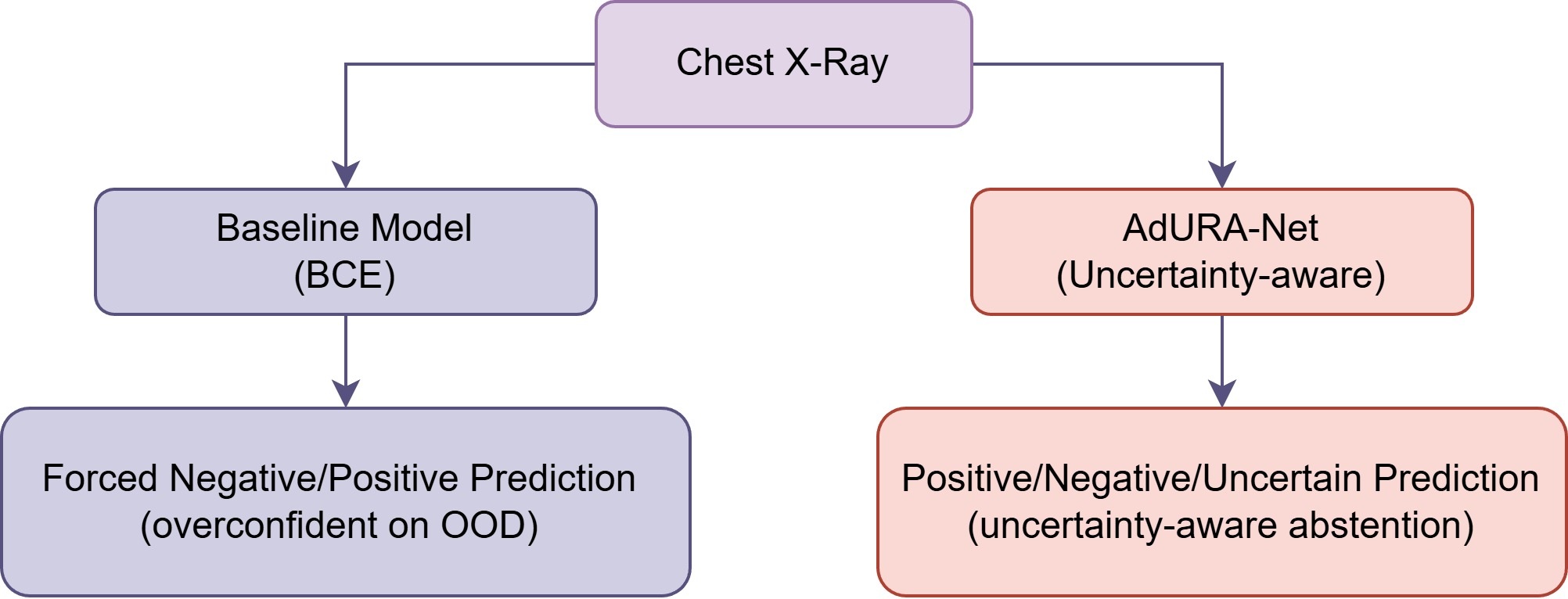}
    \caption{Conceptual overview contrasting conventional binary classification with the proposed uncertainty-aware framework, which enables three-way decisions (positive, negative, uncertain).}
    \label{fig:concept}
\end{figure}
\section{Introduction}
 In the field of clinical decision-making, uncertainty is a fundamental yet often overlooked factor. Most pathology classification models are trained under a closed-set assumption, where test samples are expected to exhibit pathological patterns similar to those observed during training for each disease. This assumption becomes problematic in real-world scenarios, where the chest X-rays may contain out-of-distribution (OOD) or ambiguous cases that are not explicitly represented during the model training. In such cases, models produce confident predictions even when sufficient evidence is not available \cite{nguyen2015deep}, which leads to potential unreliable clinical conclusions. In large-scale medical imaging datasets such as CheXpert \cite{irvin2019chexpert}, MIMIC-CXR \cite{johnson2019mimic}, annotations include positive, negative, and uncertain labels, while PadChest \cite{bustos2020padchest} incorporates uncertainty through specific categories like differential diagnoses and suboptimal technique. Making the use of these uncertainty annotations is essential for learning clinically realistic models that can capture ambiguity rather than silencing it, therefore making the model more robust and clinically reliable. However, prior approaches, including AUCM-based optimization methods \cite{yuan2021large} or hierarchical optimization-based methods \cite{pham2021interpreting}, apply label smoothing strategies that binarize the labels. This simplifies optimization, but it artificially inflates performance and removes the model's ability to differentiate ambiguous cases from the true absence of a disease. The forced binarization fails to model uncertainty explicitly. Moreover, commonly used losses such as cross-entropy are known to produce poorly calibrated and often overconfident predictions \cite{guo2017calibration,nguyen2015deep}. By prioritizing confident class discrimination, these losses drive extreme class separation in the logit space without explicitly accounting for evidence unavailability or out-of-distribution uncertainty \cite{hendrycks2016baseline}, even in cases where radiologists themselves express uncertainty. In high-risk clinical scenarios that often require multiple diagnostic tests and expert review, the ability of a model to abstain rather than overcommit is crucial. This motivates uncertainty-aware learning frameworks in which models are encouraged not only to be accurate, but also to recognize when a prediction should be abstained. Unlike prior uncertainty-estimation approaches that either quantify predictive uncertainty at inference time \cite{linmans2023predictive} or rely on stochastic variational Bayesian approximations during training \cite{lou2025uncertainty}, our method incorporates uncertainty directly into the training objective via a deterministic Dirichlet evidential formulation. While \cite{lou2025uncertainty} necessitates computationally intensive Monte Carlo sampling and uses Causal Test-Time Adaptation (CTTA) for inference stability, our approach acts as an explicit regularizer that enables calibrated abstention within a single forward pass, avoiding the need for post hoc estimation or complex test-time adaptation strategies. Experimentally, our model achieves an AUC of 0.9334 for distinguishing positive and negative cases. In addition, it attains a selective accuracy of 0.9528, indicating that when the model issues a confident prediction, {95.28\%} of those predictions are correct. Furthermore, the proposed approach achieves an uncertainty recall of 0.4675, meaning that {46.75\%} of the uncertain cases from the dataset are correctly identified as uncertain. These results show that our method maintains strong discriminative performance while learning meaningful abstention on ambiguous samples. We start by looking at related work to see how previous methods compare to our own, followed by a comprehensive overview of our methodology, which encompasses data and label handling strategies, the design of our model architecture, the training objective, and details. In our experiments section, we explain the evaluation metrics used, our main results, and provide an ablation study to analyze our approach. We conclude by discussing limitations and future research.
\section{Related Work}
Medical diagnosis models often face the challenge of encountering a vast diversity of diseases, with many trained on datasets such as CheXpert \cite{irvin2019chexpert} or MIMIC-CXR \cite{johnson2019mimic}, focus only on the diseases observed during training and struggle when exposed to out-of-distribution diseases, frequently predicting classes not encountered during training. Additionally, uncertainty labels in such datasets pose a significant challenge. The CheXpert dataset \cite{irvin2019chexpert} employs heuristic strategies U-Zero, U-One, and U-Ignore to handle uncertainty by assigning labels of 0, 1, or ignoring them during training. Pham et al. \cite{pham2021interpreting} proposed Label Smoothing Regularization (LSR) and applied these strategies in their CheXpert experiments. Similarly, AUCM maximization methods \cite{yuan2021large} adopt the same approach, thereby binarizing the learning problem and effectively ignoring the uncertainty factor. These approaches, while mitigating the risk of overfitting, inadvertently introduce biases by forcing models to learn distinct image distributions for the same class. However, these strategies remain limited in their ability to balance accuracy and robustness, particularly in complex medical imaging scenarios. Yang et al. \cite{yang2019learn} employed Bayesian approaches \cite{gal2016dropout} to leverage uncertainty labels in chest X-rays by modeling predictive uncertainty using Monte Carlo dropout. Their investigation reveals that incorporating uncertainty labels leads to higher predictive variance on uncertain cases, reducing overconfident predictions. However, sampling-based inference and indirect uncertainty estimation via predictive variance or entropy increase inference cost and do not explicitly model evidence. Furthermore, in multi-label medical images, the anatomical structures vary significantly, and previous methods rely on fixed kernels, leading to a loss of geometric information. To address these limitations, we integrate an adaptive deformable convolutional network \cite{yang2025intelligent,zhu2019deformable,dai2017deformable} with a Densenet backbone \cite{huang2017densely}, which enhances feature detection. While previous methods attempt to mitigate uncertainty, we propose a dual-head loss that combines masked binary cross-entropy with Dirichlet evidential loss \cite{sensoy2018evidential}. The masked binary cross-entropy component focuses on learning discriminative representations for the positive and negative classes, whereas the Dirichlet evidential loss leverages uncertainty labels to learn class-wise evidence, enabling the model to provide an explicit uncertainty outcome, a feature largely overlooked by existing methods. This hybrid approach enhances model robustness and reliability, preventing overconfidence in uncertain cases. By adopting an evidential framework, our approach enables deterministic single-pass inference and principled abstention without the computational overhead associated with Monte Carlo sampling, which requires multiple forward passes during inference. This allows our model to explicitly represent uncertainty and avoid forced binary decisions when sufficient evidence is lacking, while remaining computationally efficient. The final loss function is a hybrid of masked binary cross-entropy loss, Dirichlet evidential loss, offset loss, and orthogonal regularization. The offset loss helps in learning geometric information from the image, while orthogonal regularization aids in feature decoupling, as seen in multi-label image datasets, where features are often coupled and may cause model confusion during training. 
\section {Methodology}
\subsection {Data and Label Handling}
We used the CheXpert-Small dataset \cite{irvin2019chexpert},  on 5 thoracic pathologies: Cardiomegaly, Edema, Consolidation, Atelectasis, and Pleural Effusion. Each pathology is annotated with three possible labels: 1 (positive), 0 (negative), and –1 (uncertain). The official validation split of the CheXpert-Small dataset contains only binary labels and does not preserve the original uncertainty annotations. Since our objective is to model and evaluate label uncertainty explicitly, we construct a validation set by partitioning the training data, thereby retaining all three label types during evaluation. In contrast to prior works \cite{yuan2021large,pham2021interpreting}, which evaluate solely on binary labels, our evaluation protocol includes uncertain cases, enabling a more realistic assessment of uncertainty-aware learning. The dataset is split using a 96:4 ratio, with 96\% of the samples used for training and the remaining 4\% held out for validation. This results in 183,385 training samples and 7,642 validation samples, the data distribution for each class have been shown in Table \ref{tab:label_distribution}.
\begin{table}[ht]
\caption{Class-wise distribution of labels in the CheXpert-Small dataset across the training and validation splits. Labels are denoted as positive, negative, and uncertain.}
\label{tab:label_distribution}
\centering
\begin{tabular}{|l|c|c|c|c|c|c|}
\hline
\textbf{Disease} 
& \multicolumn{3}{c|}{\textbf{Train Split}} 
& \multicolumn{3}{c|}{\textbf{Validation Split}} \\
\hline
& \textbf{negative} & \textbf{positive} & \textbf{uncertain} 
& \textbf{negative} & \textbf{positive} & \textbf{uncertain} \\
\hline
Cardiomegaly      & 7,546 & 22,463 & 1,53,367 & 315 & 929 & 6398\\
\hline
Edema             & 15,276 & 47,720 & 1,20,389 & 1907 & 645 & 5090 \\
\hline
Consolidation     & 18,793 & 12,500 & 1,52,092 & 477 & 821 & 6344 \\
\hline
Atelectasis       & 998 & 28,488 & 1,53,899 & 1189 & 47 & 6406 \\
\hline
Pleural Effusion  & 24,254 & 73,788 & 85,343 & 2969 & 1046 & 3627 \\
\hline
\end{tabular}
\end{table}
\subsection{Architecture}
\begin{figure}[H]
\includegraphics[width=\textwidth]{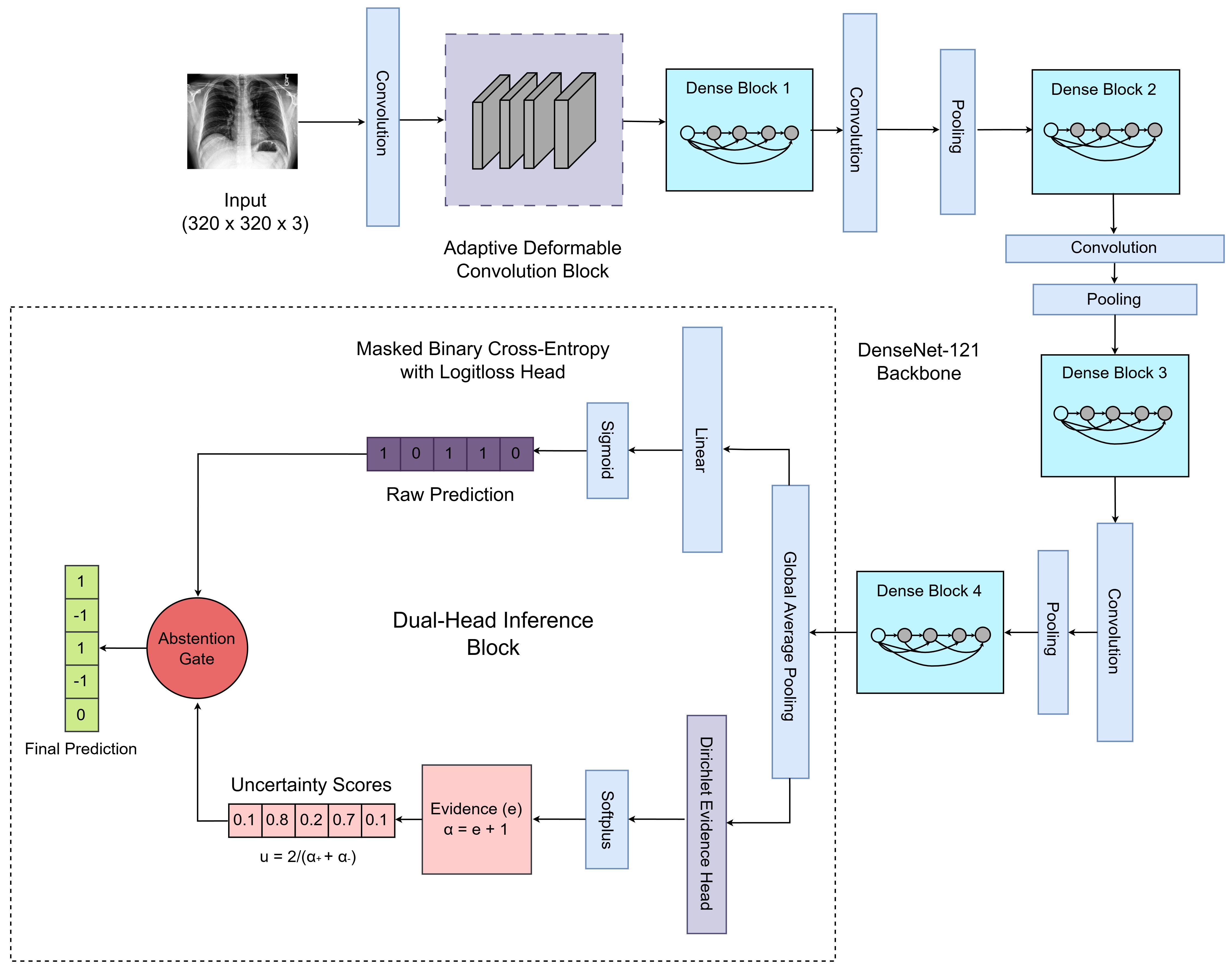}
\caption{Overview of the proposed AdURA-Net architecture. 
        The Adaptive Deformable Convolution Block enhances early geometric feature extraction. DenseNet-121 serves as the backbone for hierarchical feature propagation. A dual-head prediction module produces (1) class probabilities via a sigmoid classifier, and (2) Dirichlet Evidence $(e_i)$ for uncertainty quantification. The network is trained jointly using BCE (masked) loss, Dirichlet evidential loss, offset loss, and orthogonal regularization. During inference, the BCE head outputs the raw prediction, and the Dirichlet head outputs evidence that is used for uncertainty calculations. The abstention gate checks the uncertainty values $(u_i)$. If it is greater than the threshold $(\tau = 0.4)$, then it replaces it with $(-1)$; otherwise, it is retained.} \label{AdURAnetarch}
\end{figure}

\begin{figure}[ht]
    \centering
    \includegraphics[width=0.9\linewidth]{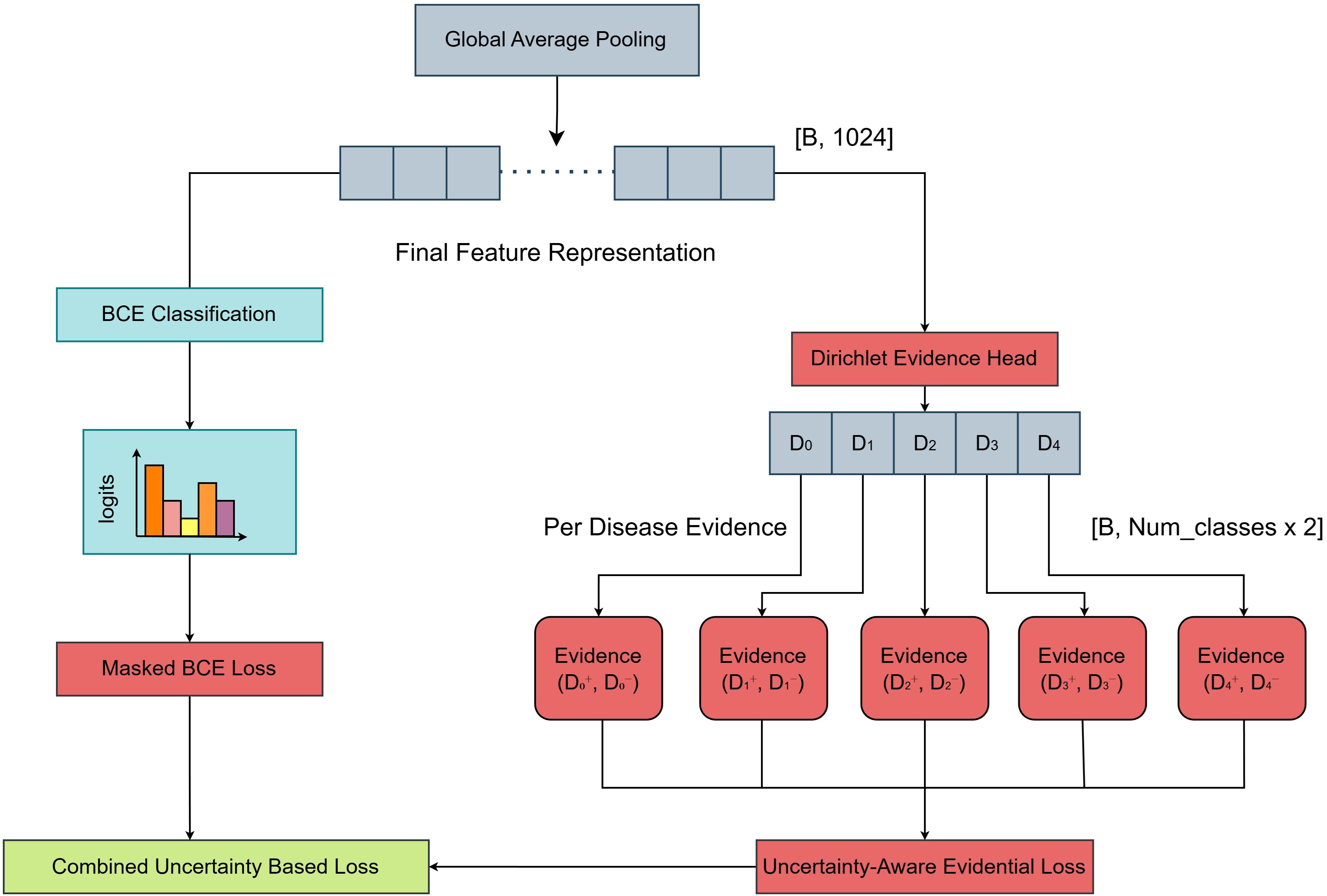}
    \caption{
    The final feature representation is processed by two parallel heads. The BCE classification head produces class-wise logits and is optimized using masked binary cross-entropy, where uncertain labels ($-1$) are masked. In parallel, the Dirichlet evidence head estimates class-wise positive and negative evidence for each disease. These evidences are supervised using an uncertainty-aware evidential loss. Both heads are jointly optimized through a combined uncertainty-based training objective. Here, $B$ denotes the batch size, and $D_0, \ldots, D_N$ represent the disease classes, each associated with positive ($D_i^{+}$) and negative ($D_i^{-}$) evidence.}
    \label{fig:loss_block}
\end{figure}

We have integrated an adaptive deformable convolution block \cite{yang2025intelligent}, which is placed immediately after the first convolution to leverage its dense preservation of spatial features. This block functions as a feature refinement module, enabling the model to learn and refine kernel displacements. This mechanism allows the model to capture the geometric structure of anatomical features. The refined feature map then flows through dense blocks, which further enhance the feature map through sustained feature propagation. The entire architecture is designed to be geometry-aware, adapting to the spatial characteristics of lesions in the input image. The final feature representation obtained after taking the global average pooling is forwarded to two parallel prediction heads: a masked binary cross-entropy (BCE) with logits head and a Dirichlet evidential head, as illustrated in Fig.~\ref{fig:loss_block}. The masked binary cross-entropy with logits head produces class logits and is optimized using only samples with definite labels. Uncertain labels ($-1$) are explicitly masked, ensuring that only positive and negative samples contribute to the discriminative loss. In parallel, the same feature representation is processed by the Dirichlet evidential head, which estimates class-wise evidence for the positive and negative classes. These evidence values are transformed into Dirichlet concentration parameters ($\alpha_+, \alpha_-$), enabling the model to quantify predictive uncertainty through the total accumulated evidence. Low evidence corresponds to high uncertainty, allowing the model to abstain from making overconfident predictions when sufficient information is not available. The evidential uncertainty loss derived from the Dirichlet head is jointly optimized with the masked binary cross-entropy loss to form the final training objective. The complete model architecture is shown in Fig.~\ref{AdURAnetarch}.
\FloatBarrier
\vspace{-5 mm}
\subsection{Training Objective}
Let $x$ denote an input chest X-ray image and 
$\mathbf{y} \in \{0,1,-1\}^C$ the corresponding multi-label annotation, 
where $0$, $1$, and $-1$ indicate negative, positive, and uncertain labels, respectively.
The network outputs class-wise logits $\mathbf{z}$, Dirichlet concentration parameters 
$\boldsymbol{\alpha}$, deformable offsets $\Delta p$, and intermediate feature maps 
$\mathbf{F}$.

\paragraph{Masked Binary Cross-Entropy Loss.}
To supervise certain labels ($y_i \in \{0,1\}$), we employ masked Binary Cross-Entropy with logits.
Uncertain labels are ignored during BCE optimization. Labels are temporarily transformed as
\begin{equation}
y_i^{\text{CE}} =
\begin{cases}
0, & \text{if } y_i = -1, \\
y_i, & \text{otherwise},
\end{cases}
\end{equation}
with a corresponding mask
\begin{equation}
m_i = \mathbb{1}[y_i \neq -1].
\end{equation}
The masked BCE loss is defined as
\begin{equation}
\mathcal{L}_{\text{BCE}} =
\frac{
\sum_{i=1}^{C} m_i \cdot \text{BCE}(z_i, y_i^{\text{CE}})
}{
\sum_{i=1}^{C} m_i
}.
\end{equation}

\paragraph{Dirichlet Evidential Loss.}
For uncertainty modeling, the network predicts Dirichlet concentration parameters
$\boldsymbol{\alpha}_i = (\alpha_i^{0}, \alpha_i^{1})$ corresponding to negative and positive evidence.
Let
\begin{equation}
S_i = \alpha_i^{0} + \alpha_i^{1}.
\end{equation}
The evidential loss is formulated as
\begin{equation}
\mathcal{L}_{\text{Dir}} =
\sum_{i: y_i = 1}
\bigl[\psi(S_i) - \psi(\alpha_i^{1})\bigr]
+
\sum_{i: y_i = 0}
\bigl[\psi(S_i) - \psi(\alpha_i^{0})\bigr]
+
\lambda_{\text{unc}}
\sum_{i: y_i = -1}
S_i,
\end{equation}
where $\psi(\cdot)$ denotes the digamma function. The final term discourages excessive evidence
accumulation for uncertain labels.

\paragraph{Orthogonal Feature Regularization.}
To encourage decorrelated channel representations, an orthogonality constraint is imposed on
intermediate feature maps. Let
$\mathbf{F} \in \mathbb{R}^{C \times HW}$ be the flattened feature representation.
The Gram matrix is computed as
\begin{equation}
\mathbf{G} = \frac{1}{HW}\mathbf{F}\mathbf{F}^\top.
\end{equation}
The orthogonality loss is defined as
\begin{equation}
\mathcal{L}_{\text{orth}} =
\lambda_{\text{orth}}
\left\lVert \mathbf{G} - \mathbf{I} \right\rVert_F^2.
\end{equation}

\paragraph{Offset Regularization Loss.}
To stabilize deformable convolution alignment, the predicted offsets $\Delta p$ are regularized
using a Huber loss:
\begin{equation}
\mathcal{L}_{\text{offset}} = \text{Huber}(\Delta p, \mathbf{0}).
\end{equation}

\paragraph{Overall Loss Function.}
The final training objective is given by
\begin{equation}
\boxed{
\mathcal{L} =
\mathcal{L}_{\text{BCE}}
+
\lambda_{\text{Dir}} \mathcal{L}_{\text{Dir}}
+
\mathcal{L}_{\text{offset}}
+
\lambda_{\text{orth}}\mathcal{L}_{\text{orth}}
}
\end{equation}

\subsection{Training Details}
Training was performed using the AdamW optimizer with momentum parameters 
$\beta_1 = 0.9$ and $\beta_2 = 0.999$. The initial learning rate was set to 
$3\times10^{-4}$ with a weight decay of $1\times10^{-5}$. We employed a cosine 
annealing learning-rate schedule with a cycle length of $T_{\max} = 100$.
For the composite loss function, the Dirichlet uncertainty weight was 
set to $\lambda_{\text{Dir}} = 0.2$, while the orthogonality regularization 
coefficient was chosen as $\lambda_{\text{orth}} = 5\times10^{-3}$. 
The Huber loss parameter was fixed at $\delta = 1.0$ to 
stabilize offset learning. The network was trained with a batch size of 16, and all DenseNet-121 backbone weights were initialized randomly without ImageNet pretraining.

\section{Experiments}
\subsection{Evaluation Metrics}

We evaluate the proposed uncertainty-aware framework using a set of complementary
metrics designed to assess both predictive performance and the model’s ability to
express uncertainty.

\paragraph{\textbf{Area Under the ROC Curve (AUC).}}
AUC is used as the primary performance metric to evaluate the model’s discriminative
capability for each disease class. For multi-label classification, AUC is computed
independently per class by considering only samples with definite labels (0 or 1).
\paragraph{\textbf{Selective Accuracy.}}
Selective accuracy measures the correctness of the model only on samples where it
makes a confident prediction. Let $C$ denote the set of samples for which the model
predicts either the positive (1) or negative (0) class, excluding abstentions ($-1$).
Selective accuracy is defined as the classification accuracy over $C$. A selective
accuracy of 95.28\% implies that 95.28\% of the model’s confident predictions are correct.
\paragraph{\textbf{Uncertainty Recall.}}
Uncertainty recall evaluates the model’s ability to correctly identify ambiguous
cases. It is defined as the proportion of ground-truth uncertain samples ($-1$)
that are predicted as uncertain by the model. \vspace{2mm} \\ 
Together, these metrics provide a comprehensive evaluation of the proposed method, capturing not only classification performance but also uncertainty calibration and principled abstention behavior.
\subsection{Main Results}
AdURA-Net was evaluated on both the 5-disease and 13-disease multi-label classification tasks. The CheXpert dataset contains 14 labels; we have considered 13 and excluded the "No Finding" labels because the purpose of our uncertainty modeling framework is to learn class-wise evidence for diseases with clinical significance. Because the "No Finding" category is a deterministic complement of the disease labels rather than an independent condition, we do not explicitly model it.  While the mean per-class AUC decreases when scaling to 13 pathologies due to increased label diversity and task difficulty, the micro-AUC remains stable, indicating that the model preserves strong overall discriminative capability as the label space expands. This behavior is consistent with prior findings by Pham et al. [15], who observed that consistently high per-class AUC is difficult to achieve with a single CNN in large-scale multi-label chest X-ray classification due to heterogeneous disease characteristics and label imbalance. For the 5-disease classification task (comprising Cardiomegaly, Edema, Consolidation, Atelectasis, and Pleural Effusion), the DenseNet-121 backbone achieves the best performance, with a Micro-AUC of 0.9334, an uncertainty recall of 0.4675, and a selective accuracy of 0.9528. This is followed by DenseNet-201, with DenseNet-161 ranking third (Table~\ref{AdURAnetbackbonecoparison5diease}). For the 13-disease classification task, DenseNet-121 again yields the strongest results, attaining a Micro-AUC of 0.9318, an uncertainty recall of 0.4245, and a selective accuracy of 0.9365 (Table~\ref{AdURAnetbackbonecoparison13diease}). We further analyze the per-class AUC for both the 5-disease and 13-disease tasks (Tables~\ref{perclassauc5disease} and~\ref{perclassauc13disease}) to assess the model’s ability to discriminate between positive and negative classes at a finer granularity. Across both settings, DenseNet-121 achieves the highest mean per-class AUC, with values of 0.8812 for the 5-disease task and 0.8335 for the 13-disease task. Overall, these results demonstrate that AdURA-Net maintains strong discriminative performance and reliable uncertainty-aware behavior, highlighting its robustness and scalability for multi-label chest X-ray classification. 
\begin{table}
\caption{Comparison of different DenseNet backbones under the proposed uncertainty-aware framework in terms of Micro-AUC, uncertainty recall, and selective accuracy for the 5 disease classification task (Cardiomegaly, Edema, Consolidation, Atelectasis, Pleural Effusion).}
\label{AdURAnetbackbonecoparison5diease}
\centering
\small
\begin{tabular}{lcccc}
\hline
\textbf{Backbones} &
\textbf{Micro-AUC} &
\textbf{Uncertainty Recall} &
\textbf{Selective Accuracy} & \\


\hline
\textbf{Densenet-121}          & \textbf{0.9334} & \textbf{0.4675} & \textbf{0.9528} \\
Densenet-161           & 0.9315 & 0.4582 & 0.9298 \\
Densenet-201           & 0.9348 & 0.4387 & 0.9315 \\
\hline
\end{tabular}
\end{table}
\begin{table}
\caption{Comparison of different DenseNet backbones under the proposed uncertainty-aware framework in terms of Micro-AUC, uncertainty recall, and selective accuracy for the 13-disease classification task (Enlarged Cardiomediastinum, Cardiomegaly, Lung Opacity, Lung Lesion, Edema, Consolidation, Pneumonia, Atelectasis, Pneumothorax, Pleural Effusion, Pleural Other, Fracture, and Support Devices).}
\label{AdURAnetbackbonecoparison13diease}
\centering
\small
\begin{tabular}{lcccc}
\hline
\textbf{Backbones} &
\textbf{Micro-AUC} &
\textbf{Uncertainty Recall} &
\textbf{Selective Accuracy} & \\
\hline
\textbf{Densenet-121}           & \textbf{0.9318} &  \textbf{0.4245} & \textbf{0.9365} \\
Densenet-161            & 0.9140 &  0.4021 & 0.9320 \\
Densenet-201            & 0.9267 & 0.4130 & 0.9310 \\

\hline
\end{tabular}
\end{table}
\begin{table}

\caption{Per-class AUC comparison across different DenseNet backbones for the 5-disease classification task (Cardiomegaly, Edema, Consolidation, Atelectasis, and Pleural Effusion).}
\label{perclassauc5disease}
\centering
\small

\begin{tabular}{lcccc}
\hline
\textbf{Class} &
Densenet-121 &
Densenet-161 &
Densenet-201 \\

\hline
Cardiomegaly            & \textbf{0.9169} & 0.9021 & 0.9121 \\
Edema                   & \textbf{0.8594} & 0.8440 & 0.8458 \\
Consolidation           & \textbf{0.9417} & 0.9302 & 0.9389 \\
Atelectasis             & \textbf{0.7397} & 0.7329 & 0.7187 \\
Pleural Effusion        & \textbf{0.9483} & 0.9312 & 0.9421 \\
\hline
\textbf{Mean AUC}     & \textbf{0.8812} & 0.8680 & 0.8715 \\
\hline
\end{tabular}
\end{table}

\begin{table}

\caption{Per-class AUC comparison across different DenseNet backbones for the 13-disease classification task (Enlarged Cardiomediastinum, Cardiomegaly, Lung Opacity, Lung Lesion, Edema, Consolidation, Pneumonia, Atelectasis, Pneumothorax, Pleural Effusion, Pleural Other, Fracture, and Support Devices).}
\label{perclassauc13disease}
\centering
\small

\begin{tabular}{lcccc}
\hline
\textbf{Class} &
\textbf{Densenet-121} &
\textbf{Densenet-161} &
\textbf{Densenet-201} &
\textbf {Ensemble} \\

\hline
Enlarged Cardiomediastinum            & 0.8712 & 0.8488 & 0.8683 & \textbf{0.8751}\\
Cardiomegaly                          & 0.9320 & 0.9158 & 0.9331 & \textbf{0.9489}\\ 
Lung Opacity                          & 0.8266 & \textbf{0.8732} & 0.8578 & 0.8395\\
Lung Lesion                           & 0.8156 & 0.8187 & 0.8054 & \textbf{0.8824}\\
Edema                                 & 0.8776 & 0.8566 & \textbf{0.8872} & 0.8780\\
Consolidation                         & 0.9426 & 0.9332 & 0.9441 & \textbf{0.9504}\\
Pneumonia                             & 0.7840 & 0.7163 & 0.7868 & \textbf{0.7888} \\
Atelectasis                           & 0.7784 & \textbf{0.7844} & 0.7690 & 0.6962\\
Pneumothorax                          & 0.7663 & 0.7526 & \textbf{0.7750} & 0.7717\\
Pleural Effusion                      & 0.9482 & 0.9463 & \textbf{0.9505} & 0.9485\\
Pleural Other                         & 0.7981 & 0.7875 & 0.7480 & \textbf{0.8777}\\
Fracture                              & \textbf{0.7820} &  0.7536 & 0.7449 & 0.7525\\
Support Devices                       & 0.7129 & \textbf{0.7150} & 0.7036 & 0.7127\\
\hline
\textbf{Mean AUC}                     & 0.8335 & 0.8232 & 0.8287 & \textbf{0.8402}\\
\hline
\end{tabular}
\end{table}

From the above results, we observe that DenseNet-121 consistently outperforms deeper backbone architectures for the 5-disease classification task under uncertainty-aware training. One contributing factor is the reduced effective training set size caused by masking uncertain labels, as a substantial portion of the CheXpert annotations are marked as uncertain. This reduction in available supervision disproportionately affects larger architectures, which typically require more data to fully exploit their increased capacity, leading to diminished performance gains. For the 13-disease classification task, we note that for certain disease categories, such as Edema, Pleural Effusion, etc., deeper backbones exhibit improved per-class performance. This behavior can be attributed to the increased semantic diversity and complexity introduced by the larger label set, where higher-dimensional feature representations may be beneficial for modeling specific pathologies. Motivated by this complementary behavior across backbone depths, we further evaluated a lightweight ensemble formed by averaging the logits of DenseNet-121, DenseNet-161, and DenseNet-201. This simple aggregation yielded slightly improved overall performance, suggesting that combining representations from multiple capacity regimes can better capture heterogeneous disease characteristics. These observations suggest that while a moderately deep backbone provides the best overall balance between robustness and capacity under uncertain supervision, deeper architectures can offer localized advantages for particular diseases in more complex classification settings.
\subsection{Ablation Studies}
\subsubsection{Analysis of the Adaptive Deformable Convolution Block}
To assess the impact of the employed Adaptive Deformable Convolution Block, we first conducted an ablation study using the CheXpert-small dataset \cite{irvin2019chexpert}. Using the U-Zero and U-One evaluation protocols, the baseline DenseNet-121 achieved an average AUC of 0.8741 and 0.8828 respectively, whereas the addition of the adaptive deformable convolution block improved the AUC to 0.8902 for the U-Zero setup and 0.8856 for the U-One setup (Table \ref{perclassauc_adcbd121}), demonstrating the benefit of geometry-aware feature refinement. We also evaluated multiple DenseNet backbone variants (Table \ref{perclassauc_adcb161201} and Table \ref{perclassauc_adcb161201_uone}) and observed that DenseNet-201 achieved the best performance, with an average AUC of 0.8930 and 0.8918 for U-Zero and U-One setup respectively, compared to other DenseNet models \cite{huang2017densely} for the 5 disease classification task.
\begin{table}
\caption{Per-class AUC comparison for baseline DenseNet-121 and the employed adaptive deformable model under U-Zero and U-One evaluation settings for the 5 disease (Cardiomegaly, Edema, Consolidation, Atelectasis and Pleural Effusion) classification task.}
\label{perclassauc_adcbd121}
\centering
\small
\begin{tabular}{lcccc}
\hline
\textbf{Class} &
\textbf{Base U-Zero} &
\textbf{Base U-One} &
\textbf{U-Zero(Ours)} &
\textbf{U-One(Ours)} \\
\hline
Cardiomegaly            & 0.8474 & 0.8264 & \textbf{0.8482} & \textbf{0.8714} \\
Edema                   & 0.8647 & 0.8957 & \textbf{0.8997} & \textbf{0.9137} \\
Consolidation           & 0.9090 & 0.8994 & \textbf{0.9380} & \textbf{0.8781} \\
Atelectasis             & 0.8448 & \textbf{0.8710} & 0.8425 & 0.8215 \\
Pleural Effusion        & 0.9047 & 0.9219 & \textbf{0.9226} & \textbf{0.9452} \\
\hline
\textbf{Mean AUC}     & 0.8741 & 0.8828 & \textbf{0.8902} & \textbf{0.8856} \\
\hline
\end{tabular}
\end{table}
\begin{table}
\caption{Per-class AUC comparison for the adaptive deformable convolution block integrated with different Densenet backbones under U-Zero setting for the 5 disease (Cardiomegaly, Edema, Consolidation, Atelectasis, and Pleural Effusion) classification task.}
\label{perclassauc_adcb161201}
\centering
\small
\begin{tabular}{lcccccc}
\hline
\textbf{Backbone} &
\textbf{Cardiomegaly} &
\textbf{Edema} &
\textbf{Consolidation} &
\textbf{Atelectasis} &
\textbf{Pleural Effusion}&
\textbf{Mean AUC}\\
\hline
densenet-161            & 0.8543 & 0.9117 & 0.9360 & 0.8125 & 0.9288 & \textbf{0.8886} \\
densenet-201                   & 0.8420 & 0.9264 & 0.9167 & 0.8460 & 0.9343 & \textbf{0.8930} \\
\hline
\end{tabular}
\end{table}
\begin{table}
\caption{Per-class AUC comparison for the adaptive deformable convolution block integrated with different Densenet backbones under U-One setting for the 5 disease (Cardiomegaly, Edema, Consolidation, Atelectasis, and Pleural Effusion) classification task.}
\label{perclassauc_adcb161201_uone}
\centering
\small
\begin{tabular}{lcccccc}
\hline
\textbf{Backbone} &
\textbf{Cardiomegaly} &
\textbf{Edema} &
\textbf{Consolidation} &
\textbf{Atelectasis} &
\textbf{Pleural Effusion}&
\textbf{Mean AUC}\\
\hline
densenet-161            & 0.8545 & 0.9012 & 0.9061 & 0.8474 & 0.9214 & \textbf{0.8861} \\
densenet-201                   & 0.8537 & 0.9199 & 0.9023 & 0.8565 & 0.9267 & \textbf{0.8918} \\
\hline
\end{tabular}
\end{table}
We observe that larger backbones, such as DenseNet-201, achieve competitive performance in certain settings. The contributing factors can be the increased dimensionality of the final feature representation (1920 channels for DenseNet-201 compared to 1024 channels for DenseNet-121), which provides greater representational capacity for modeling complex disease patterns when sufficient training samples are available under the U-Zero or U-One setup, which was not the case under the uncertainty-aware training. However, excessively large feature representations, as in DenseNet-161 with 2208 channels, may introduce redundant or overly complex feature interactions, leading to diminishing returns. These findings suggest that complex diseases can benefit from increased feature dimensionality; an appropriate balance between representational capacity and architectural complexity is essential.
\subsubsection{Confidence Comparison of Uncertainty model Vs Base model}
\begin{figure}[H]
    \centering
    \includegraphics[width=0.9\linewidth]{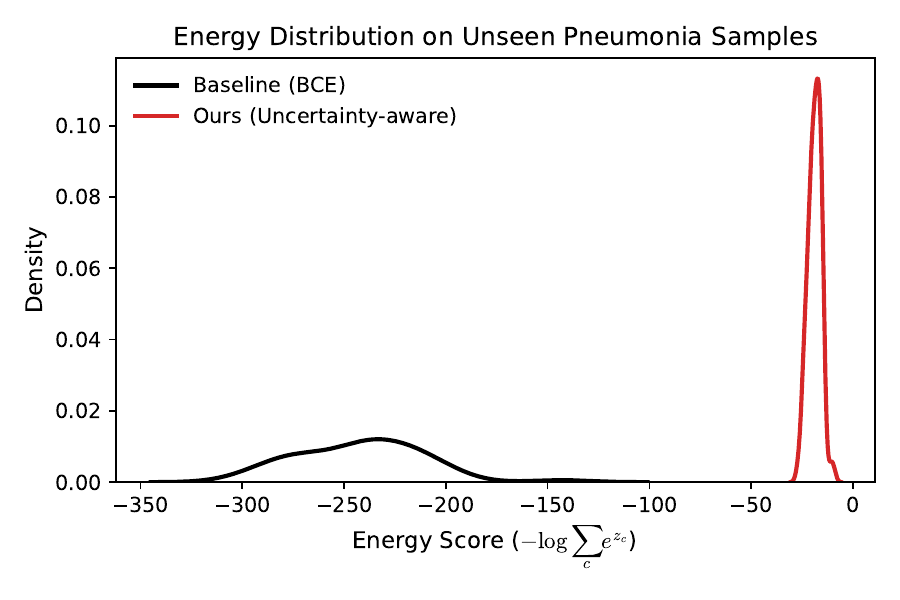}
    \caption{Energy distribution on unseen pneumonia samples. Lower energy indicates higher confidence. The uncertainty-aware model shifts the distribution toward higher energy values, reducing overconfident predictions.}
    \label{fig:energypenumonia}
\end{figure}

\begin{figure}[H]
    \centering
    \includegraphics[width=0.9\linewidth]{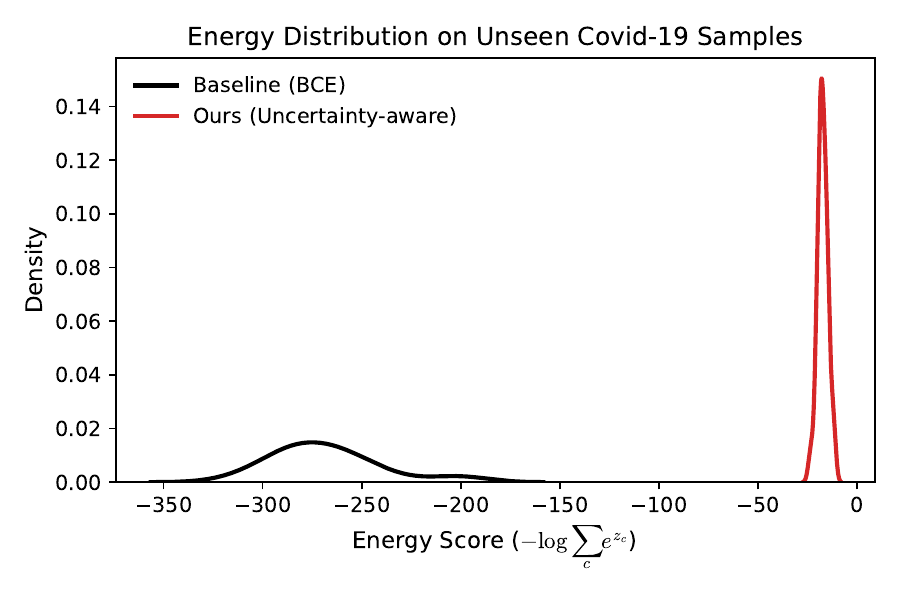}
    \caption{Energy distribution on unseen Covid-19 samples. Lower energy indicates higher confidence. The uncertainty-aware model shifts the distribution toward higher energy values, reducing overconfident predictions.}
    \label{fig:energycovid19}
\end{figure}

To evaluate the effectiveness of the proposed uncertainty-aware model, we compare its predictive confidence against the uncertainty-unaware baseline Densenet-121 model on out-of-distribution chest X-ray samples corresponding to Pneumonia \cite{kermany2018labeled,shih2019augmenting} and COVID-19 \cite{jamdade2020covid}, obtained from an external dataset. As shown in the energy distribution plots in Fig\ref{fig:energypenumonia}. and Fig\ref{fig:energycovid19}., The uncertainty-aware model produces energy values that are closer to zero (i.e., less negative), indicating higher predictive uncertainty. This behavior is consistent with the formulation of energy-based uncertainty, where higher energy corresponds to lower confidence and is expected for previously unseen or out-of-distribution disease patterns. In contrast, the baseline uncertainty-unaware model assigns highly negative energy values to the same samples, reflecting overconfident predictions despite the lack of supporting evidence. These results align with prior work on energy-based out-of-distribution detection, which demonstrates that energy scores provide a more reliable measure of uncertainty than softmax confidence by avoiding spurious high-confidence predictions on unseen data \cite{liu2020energy}. This comparison demonstrates that the proposed approach effectively mitigates overconfident predictions and provides a more reliable uncertainty signal when confronted with unfamiliar clinical conditions.
\subsubsection{Effect of Uncertainty Modeling on Binary Classification Performance}
\begin{figure}
    \centering
    \includegraphics[width=\textwidth]{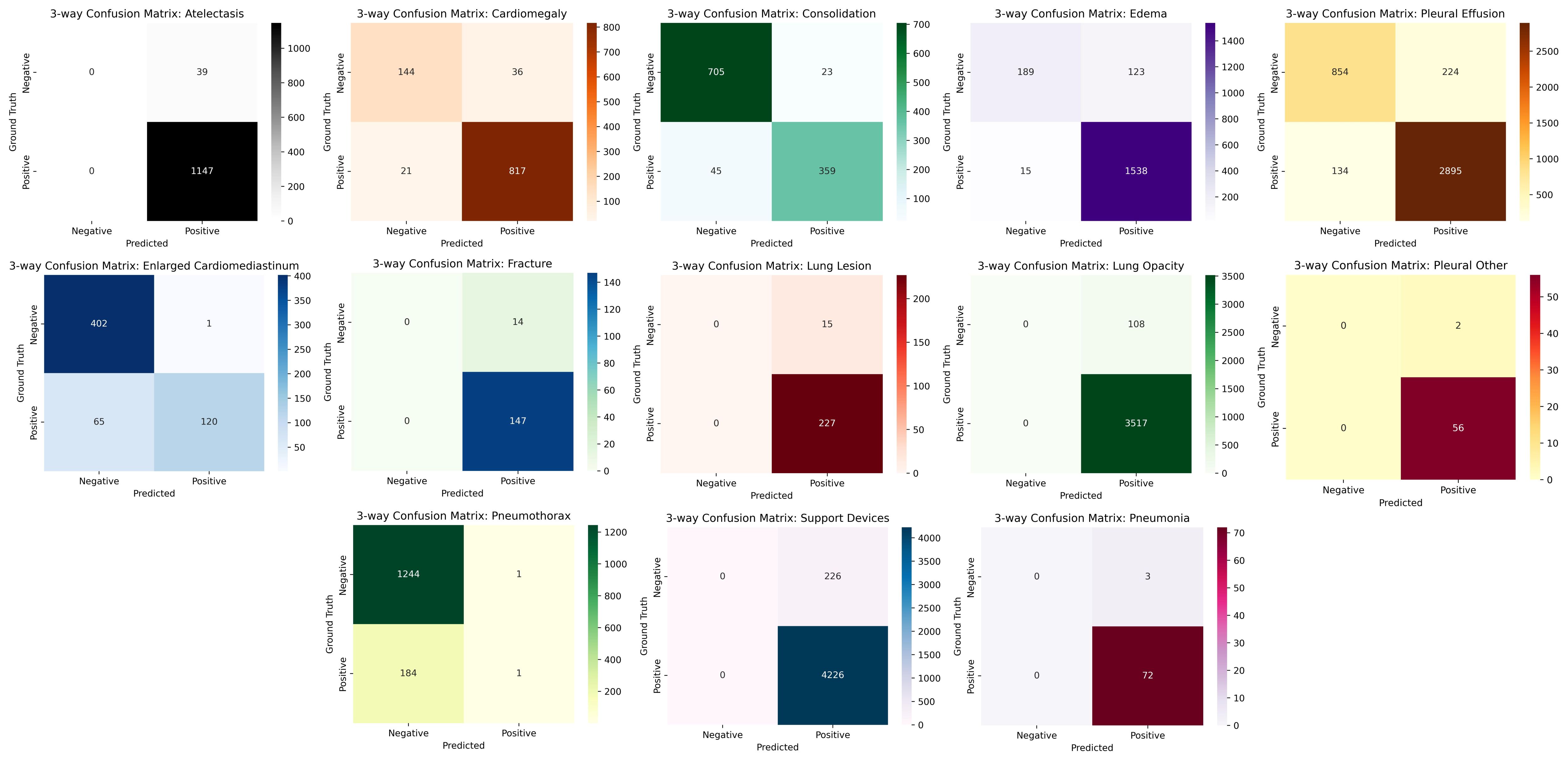}
    \caption{
    Per-class binary confusion matrices (positive vs.\ negative) for the proposed uncertainty-aware model.
    This analysis evaluates whether incorporating uncertainty modeling affects standard class discrimination,
    demonstrating that positive and negative prediction performance is preserved across all disease categories.
    }
    \label{fig:binary_confusion_ablation}
\end{figure}

We further analyze the impact of uncertainty modeling on standard classification by visualizing per-class binary confusion matrices. The results in  Fig.\ref{fig:binary_confusion_ablation}. indicate that incorporating evidential uncertainty does not degrade positive–negative discrimination across disease categories. Note that, because the dataset is imbalanced, some disease categories contain very few negative cases, and a few have none; this scarcity is especially pronounced for certain diseases, as shown in Fig.~\ref{fig:binary_confusion_ablation}.

\section{Limitations}
We have used the uncertainty labels from the CheXpert-small dataset \cite{irvin2019chexpert}. However, these labels are not always accurate, as they often reflect cases where the radiologist was uncertain, but they are also mixed with instances where the label extractor failed to extract the correct label or due to the ambiguity in the report \cite{irvin2019chexpert}. In such cases, the uncertain labels may include both positive and negative class predictions. This creates a challenge for the model, as it may output a positive or negative class even when the label is uncertain. To address this, we need to incorporate uncertainty labels alongside the positive and negative classes, which are actually uncertain cases. This approach helps the model learn not only to be confident but also to recognize when it should abstain from making predictions. We observe that improper scaling of $\lambda_{\text{Dir}}$ may lead to numerical instability, highlighting the importance of careful hyperparameter selection when combining evidential objectives with discriminative losses. Due to computational constraints, all images are resized to a resolution of 320 × 320. We note that training at higher spatial resolutions may further improve performance. Future work could expand the evidential framework to include logical or derived labels like "No Finding," since we do not include this label during modeling. While the results are encouraging, further evaluation on additional medical imaging datasets such as MIMIC-CXR \cite{johnson2019mimic} that explicitly contain uncertainty annotations is necessary to assess the generalizability of the proposed approach.

\printbibliography[
heading=bibintoc,
title={References}
]

\end{document}